\documentclass[letterpaper, 10 pt, conference]{ieeeconf}  

\IEEEoverridecommandlockouts                              

\usepackage{graphics} 
\usepackage{epsfig} 
\usepackage{mathptmx} 
\usepackage{times} 
\usepackage{amsmath} 
\usepackage{amssymb}  
\usepackage{subcaption}
\usepackage{multirow}
\usepackage{url}
\title{\LARGE \bf
Bi-VLA: Bilateral Control-Based Imitation Learning via Vision-Language Fusion for Action Generation
}

\author{
Masato Kobayashi$^{1,2,3\dag*}$, Thanpimon Buamanee$^{2\dag}$
\thanks{
${\dag}$ Equal Contribution,
$^{1}$ D3 Center, The University of Osaka,$^{2}$ Graduate School of Information Science and Technology, The University of Osaka, $^{3}$ Graduate School of Maritime Sciences, Kobe University, * corresponding author: kobayashi.masato.cmc@osaka-u.ac.jp
}
}

\begin{document}

\maketitle
\thispagestyle{empty}
\pagestyle{empty}

\begin{abstract}
We propose Bilateral Control-Based Imitation Learning via Vision-Language Fusion for Action Generation (Bi-VLA), a novel framework that extends bilateral control-based imitation learning to handle more than one task within a single model. Conventional bilateral control methods exploit joint angle, velocity, torque, and vision for precise manipulation but require task-specific models, limiting their generality.
Bi-VLA overcomes this limitation by utilizing robot joint angle, velocity, and torque data from leader-follower bilateral control with visual features and natural language instructions through SigLIP and FiLM-based fusion.
We validated Bi-VLA on two task types: one requiring supplementary language cues and another distinguishable solely by vision. Real-robot experiments showed that Bi-VLA successfully interprets vision-language combinations and improves task success rates compared to conventional bilateral control-based imitation learning.
Our Bi-VLA addresses the single-task limitation of prior bilateral approaches and provides empirical evidence that combining vision and language significantly enhances versatility.
Experimental results validate the effectiveness of Bi-VLA in real-world tasks.
For additional material, please visit the website: \url{https://mertcookimg.github.io/bi-vla/}

\end{abstract}

\section{INTRODUCTION}
Robotic manipulation is increasingly important in human-centered applications such as cooking, eldercare, and interactive service robots~\cite{cooking2024Sakib,cooking2025Verghese,service2024Škerlj,service2025Yang}.
Unlike traditional industrial robots that excel at repetitive and pre-programmed routines, service and collaborative robots must adapt to dynamic environments and interact with objects of diverse shapes, sizes, and material properties~\cite{in2023He,in2023Huang}. Achieving such adaptability requires learning frameworks capable of acquiring human-like manipulation strategies~\cite{firoozi2023foundationmodelsroboticsapplications}.

Imitation learning (IL) has emerged as a promising approach for transferring human manipulation skills directly to robots~\cite{tsuji2025surveyimitationlearningcontactrich}.
Leader-follower teleoperation has become a common pipeline for collecting demonstrations. For example, ALOHA and Mobile ALOHA use position-based unilateral control to gather diverse datasets that enable a wide range of manipulation~\cite{fu2024mobilealohalearningbimanual, aloha2team2024aloha2enhancedlowcost}.
Although effective for kinematics-driven tasks, such unilateral control omits force feedback, which limits robustness in contact-rich interactions.

Bilateral control-based imitation learning addresses these limitations by exchanging both position and force information between the demonstrator and robot~\cite{adachi2018imitation, sakaino2022imitation}.
Bilateral control allows demonstrators to feel contact forces directly, yielding richer demonstrations and improving generalization across objects with different hardness and weights.
\begin{figure}[t]
\centering
\includegraphics[keepaspectratio, width=0.95\linewidth]{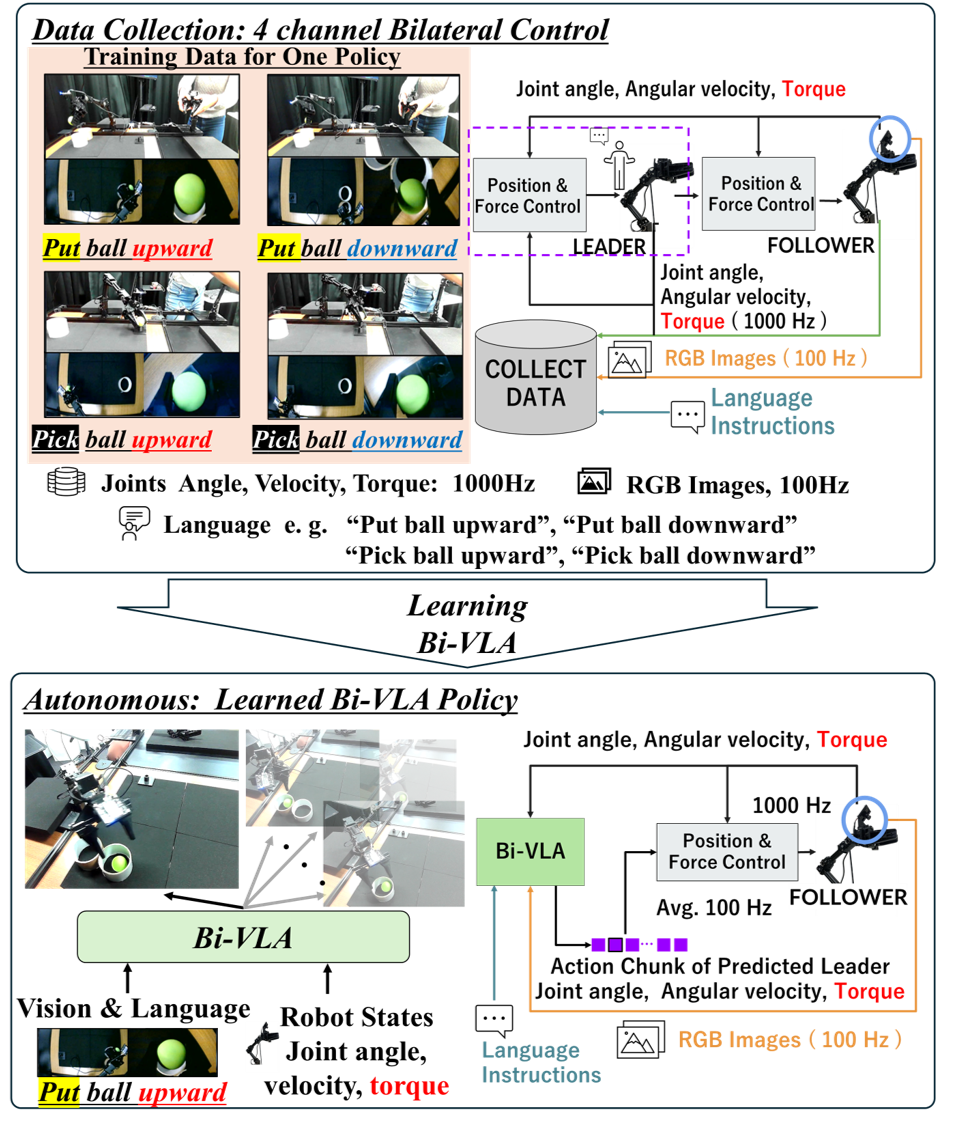}
\caption{Bi-VLA: Bilateral Control-Based Imitation Learning via Vision-Language Fusion for Action Generation}
\label{fig:teser}
\end{figure}

Building on this foundation, recent methods have explored combining bilateral control with modern architectures.
Bilateral Control-Based Imitation Learning via Action Chunking with Transformers (Bi-ACT)~\cite{buamanee2024bi} integrates bilateral control with visual observations via Transformers, yielding improved manipulation accuracy.
However, it remains restricted to single-task settings, limiting its practicality in dynamic environments where multiple tasks must be handled seamlessly.
In parallel, Bilateral Control-Based Imitation Learning via Natural Language and Action Chunking with Transformers (Bi-LAT)~\cite{kobayashi2025bilat} introduced natural language instructions into bilateral control-based imitation learning, demonstrating effective force modulation in manipulation. 
While this work highlights the promise of language integration, it focuses on regulating applied force and does not address the challenge of enabling a single model to adapt to multiple task contexts.

In this paper, we propose Bilateral Control-Based Imitation Learning via Vision-Language Fusion for Action Generation (Bi-VLA), a novel framework that unifies robot joint angle, velocity, and torque data from bilateral control with visual and language features, as shown in Fig.~\ref{fig:teser}.
By leveraging SigLIP-based text embeddings~\cite{zhai2023sigmoid} and FiLM-based EfficientNet feature fusion~\cite{perez2017filmvisualreasoninggeneral}, Bi-VLA learns a shared representation of vision and language aligned with robot state information. Unlike prior bilateral frameworks, Bi-VLA is designed to handle multiple tasks within a single model, enabling flexible task switching without retraining or manual model selection.

The main contributions of this paper are summarized as follows:
\begin{itemize}
    \item We propose Bi-VLA, the bilateral control-based imitation learning framework that fuses vision and language features into a unified representation.
    \item We demonstrate that Bi-VLA enables a single model to perform multiple tasks, overcoming the single-task limitation inherent in prior bilateral control-based imitation learning approaches.
    \item We validate Bi-VLA through real-robot experiments on two distinct tasks, showing improved performance and adaptability compared to conventional methods.
\end{itemize}

\section{RELATED WORK}

\subsection{Vision-Language Information for Robotic Manipulation }
Vision-language models (VLMs) such as Contrastive Language-Image Pre-Training (CLIP) have advanced vision and language-based manipulation~\cite{vlm2024Gao, zhou2025bridginglanguageactionsurvey}.
Representative systems include CLIPort~\cite{shridhar2021cliport}, which exploits pre-trained CLIP features for instruction-following pick-and-place, and PerAct~\cite{shridhar2022perceiveractormultitasktransformerrobotic}, which integrates language, vision, and action via 3D attention to acquire more complex policies.
Beyond CLIP pipelines, modular stacks that pair a language encoder(e.g., DistilBERT~\cite{sanh2019distilbert}) with a visual backbone(e.g., EfficientNet~\cite{tan2020efficientnetrethinkingmodelscaling}) and fuse them using FiLM~\cite{perez2017filmvisualreasoninggeneral} have enabled flexible instruction grounding; prior studies also indicate that the choice of language encoder substantially affects grounding fidelity and task success~\cite{rt12022arxiv, shi2024yellrobotimprovingonthefly}.
However, most of these frameworks rely on unilateral teleoperation and primarily kinematic control, neglecting force cues that are crucial for robust, contact-rich behaviors and human-like compliance~\cite{shi2024yellrobotimprovingonthefly}.

Our Bi-VLA method addresses these challenges by taking the previous model~\cite{shi2024yellrobotimprovingonthefly} as a base, replacing the text encoder from DistilBERT to SigLIP, maintaining EfficientNet for visual processing, and fusing the modalities with FiLM.
The results are then embedded into a four-channel bilateral control pipeline that leverages both position and force information.
This design preserves the generality of VLMs while restoring contact sensitivity, enabling a single policy to switch between tasks through language cues without sacrificing robustness in contact.

\subsection{Bilateral Control-Based Imitation Learning}

Bilateral control is a leader-follower system that combines position control and force control based on position tracking and action-reaction law.
Through the leader robot, the operator can command the follower robot’s motion while simultaneously perceiving in real time the contact forces experienced by the follower.

Bilateral control-based imitation learning has demonstrated the importance of force feedback for generalization across objects and materials.
Using recurrent networks and  Long Short Term Memory(LSTM), such approaches have successfully reproduced force-dependent skills such as ruler-based drawing, surface wiping, and precise slicing~\cite{adachi2018imitation, sakaino2022imitation, hayashi2022independently}.
In addition, data augmentation methods that accommodate variable execution speeds have also been proposed~\cite{Masuya19052025}.

More recently, transformer-based architectures and emerging sequence models such as Mamba have been incorporated into bilateral control-based imitation learning~\cite{buamanee2024bi,tsuji2025manba}.
For instance, Bi-ACT extends the Action Chunking with Transformers (ACT), originally used in ALOHA and Mobile ALOHA, by integrating position, force, and visual information, thereby achieving strong generalization across diverse objects~\cite{buamanee2024bi, zhao2023learning}.
ALPHA-$\alpha$ provides a low-cost bimanual robot arm platform for collecting position and force data via bilateral control~\cite{kobayashi2025alpha}, while DABI introduces a data augmentation technique that integrates high-frequency robot signals with low-frequency visual data via downsampling~\cite{kobayashi2024dabievaluationdataaugmentation}. Furthermore, Bi-LAT incorporates natural language instructions to modulate force during manipulation, suggesting the potential of language as a conditioning signal for force control~\cite{kobayashi2025bilat}.

Our Bi-VLA method introduces a novel framework that integrates both visual and linguistic modalities into bilateral control-based imitation learning.
We fuse SigLIP-based text embeddings and EfficientNet visual features through FiLM.
These multimodal features are then combined with position and force information and embedded into a transformer-driven conditional variational autoencoder (CVAE).
Unlike prior approaches, which either trained task-specific policies or used language solely for force modulation, the proposed Bi-VLA enables a single policy to generalize across multiple tasks by leveraging vision and language information.

\begin{figure}[t]
\centering
\includegraphics[keepaspectratio, width=0.9\linewidth]{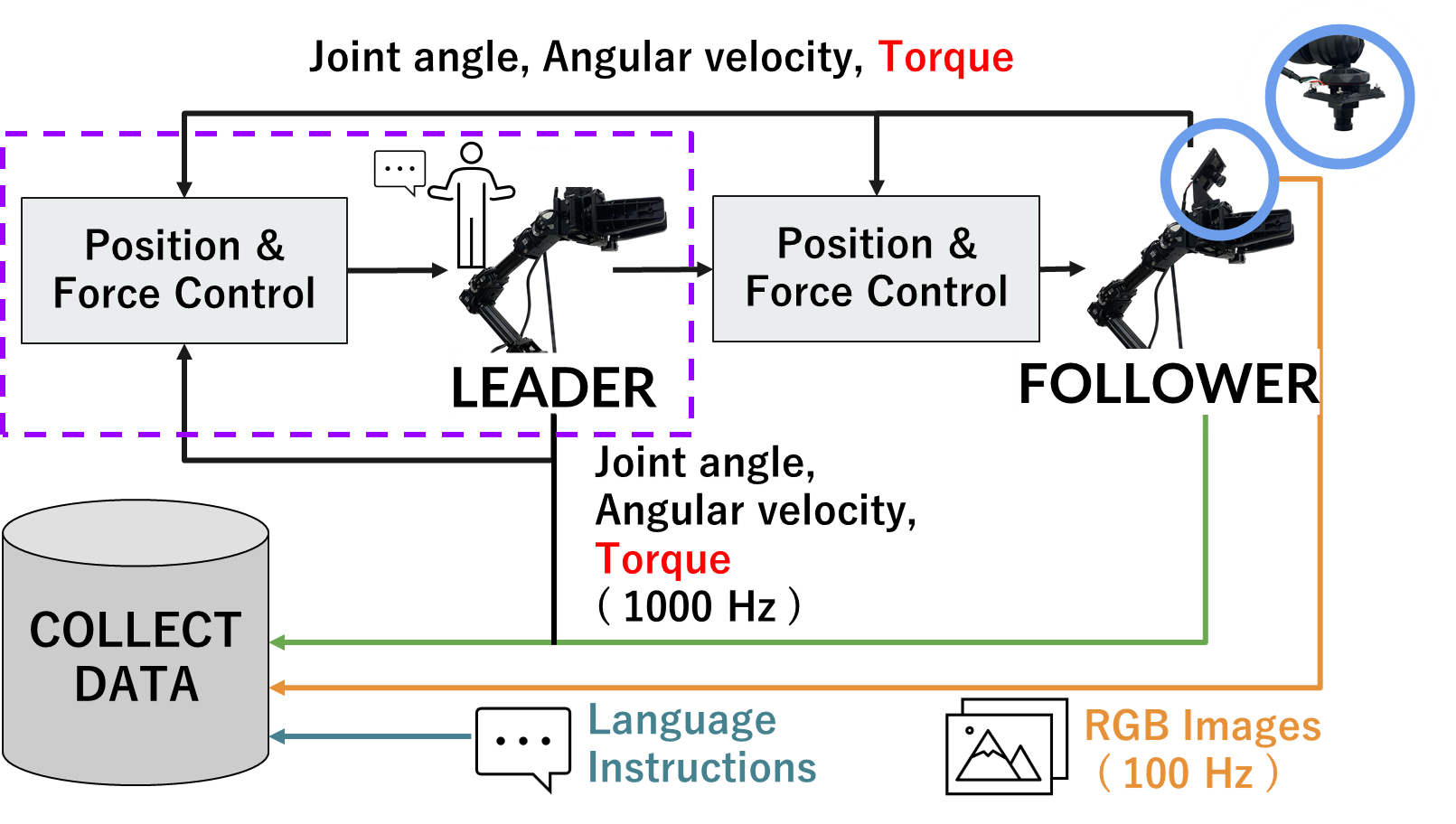}
\caption{Bi-VLA: Data Collection (Bilateral Control)}
\label{fig:bi-vla_datacollection}
\end{figure}

\begin{figure*}[t]
\centering
\includegraphics[keepaspectratio, width=0.85\linewidth]{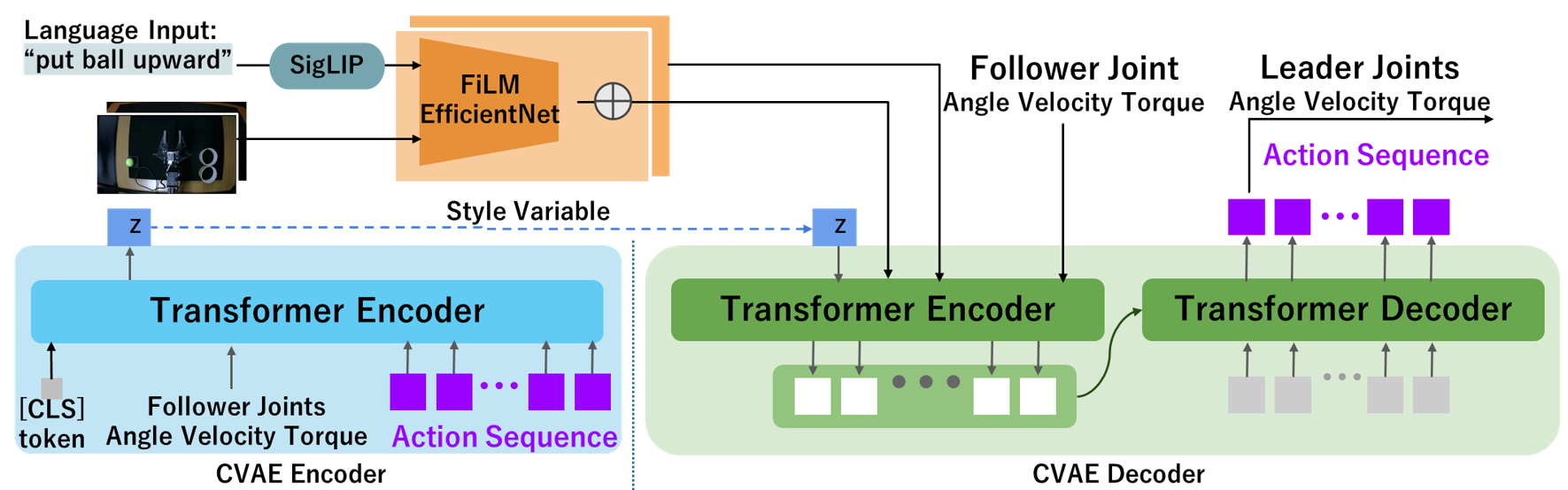}
\caption{Bi-VLA: Learning Model}
\label{fig:bi-vla-model}
\end{figure*}

\section{Bi-VLA: Bilateral Control-Based Imitation Learning via Vision-Language Fusion for Action Generation}
\subsection{Overview}\label{sec:bi-vla_overview}
Previous bilateral control-based imitation learning approaches, such as Bi-ACT, combined positional and force information with visual inputs to achieve adaptive robot control. However, these methods were restricted to single-task execution. 

To address this limitation, we introduce Bi-VLA, which extends the bilateral control-based imitation learning framework by incorporating a language encoder and vision-language fusion. 
By integrating natural language instructions with visual observations and robot joint data (angle, velocity, and torque), Bi-VLA enables a single model to flexibly perform multiple tasks conditioned on user-provided commands. 

\subsection{Data Collection}\label{sec:bi-vla_datacollection}
Data were collected using a four-channel bilateral control system, in which the leader and follower robots continuously exchanged position and torque information to ensure coordinated behavior. 
The bilateral control law is given by:
\begin{equation}
\theta_l - \theta_f = 0,
\label{eq:position}
\end{equation}
\begin{equation}
\tau_l + \tau_f = 0,
\label{eq:force}
\end{equation}
where $\theta$ and $\tau$ denote the joint angle and torque, respectively, with subscripts $l$ and $f$ indicating the leader and follower. 
Joint angles were obtained from encoders, while external torques were estimated using a disturbance observer (DOB)~\cite{ohnishi1996motion} and a reaction force observer (RFOB)~\cite{murakami2002torque}, thereby eliminating the need for dedicated force/torque sensors.

In addition to robot joint data (angle, velocity, and torque) and visual inputs, natural language instructions were recorded during bilateral control as shown in Fig.~\ref{fig:bi-vla_datacollection}. 
The operator manipulated the leader robot to control the follower while providing language instructions describing the intended actions. 
These instructions were temporally aligned with robot states and RGB images, yielding a multimodal dataset for training.

\begin{figure}[t]
\centering
\includegraphics[keepaspectratio, width=0.95\linewidth]{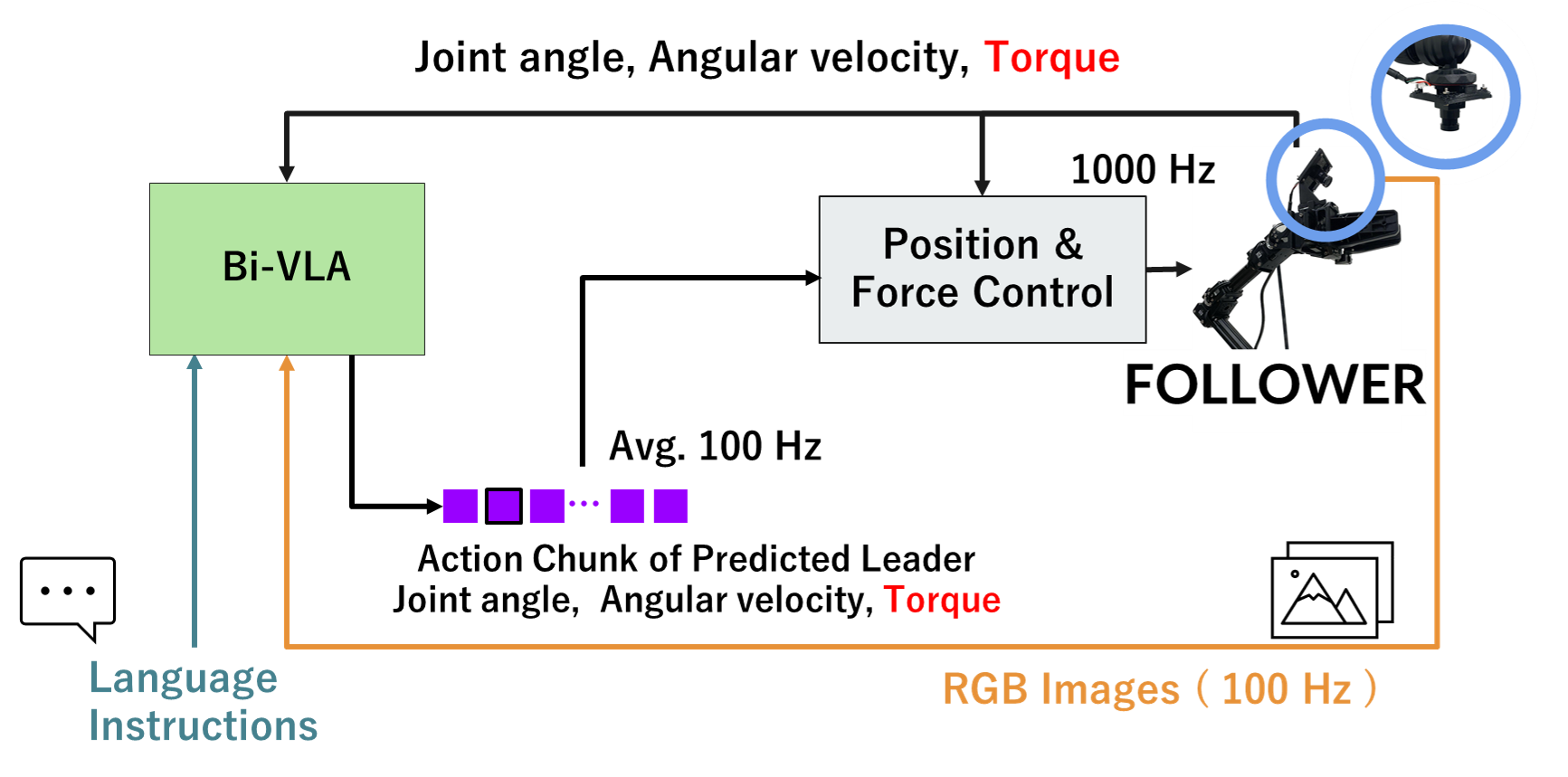}
\caption{Bi-VLA: Inference}
\label{fig:bi-vla_inference}
\end{figure}

\begin{figure*}[t] 
    \centering 
    \includegraphics[keepaspectratio, width=0.95\linewidth]{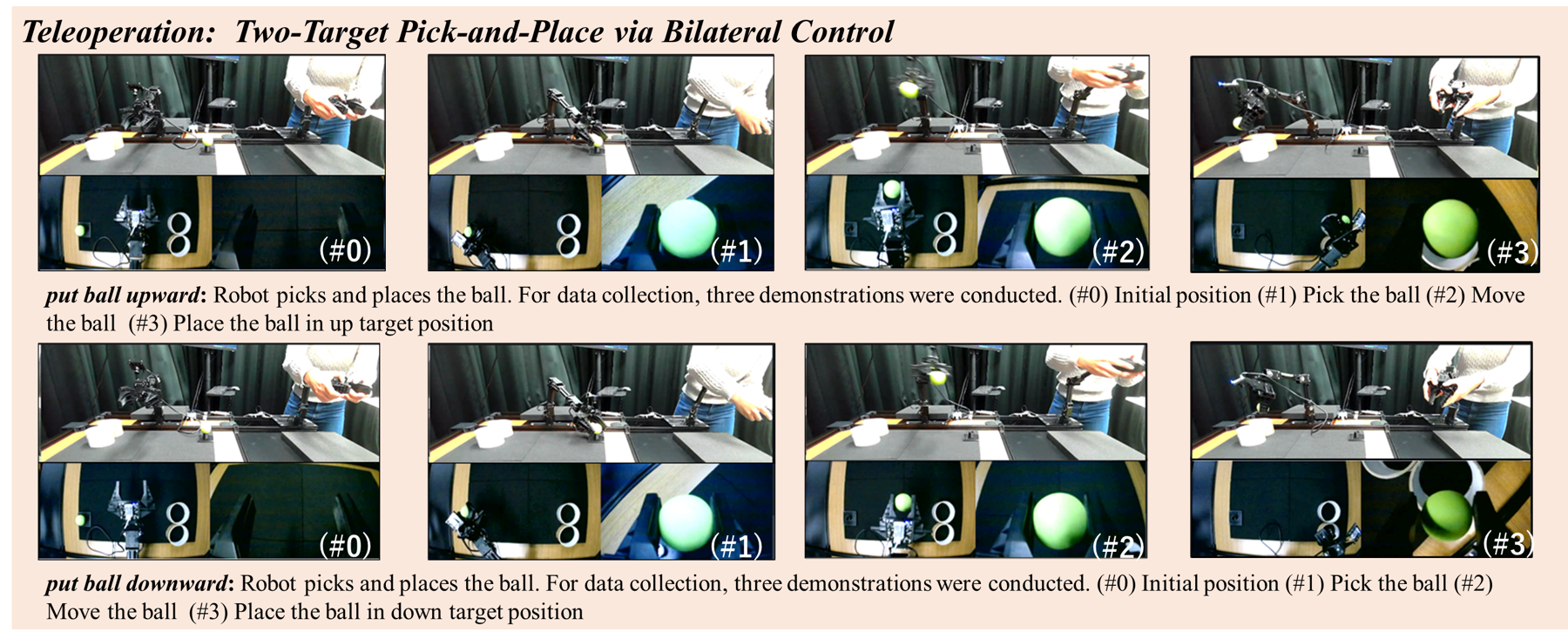} 
    \caption{Data Collection of Two-Target Pick-and-Place Task} 
    \label{fig:ex1_tele} 
\end{figure*}

\begin{figure*}[t] 
    \centering 
    \includegraphics[keepaspectratio, width=0.95\linewidth]{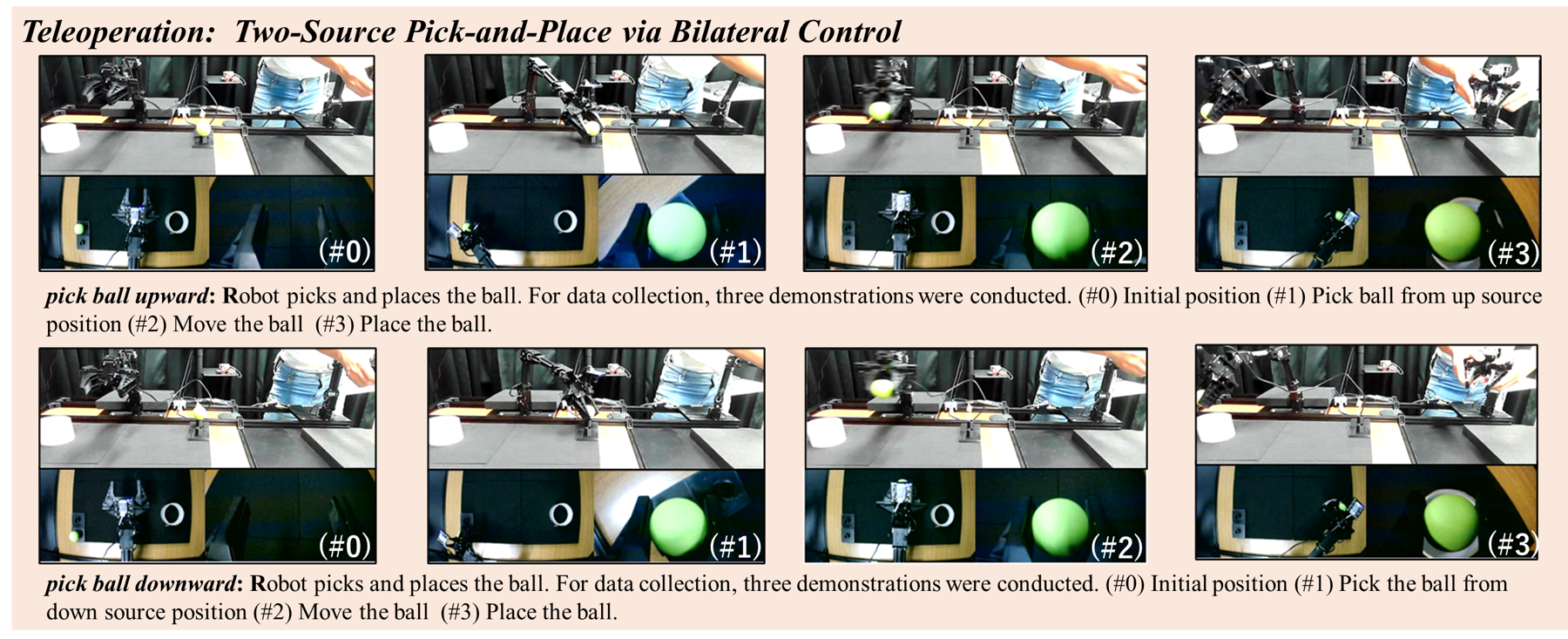} 
    \caption{Data Collection of Two-Source Pick-and-Place Task} 
    \label{fig:ex2_tele} 
\end{figure*}

\subsection{Learning Model}\label{sec:bi-vla_model}
As shown in Fig.~\ref{fig:bi-vla-model}, Bi-VLA is implemented using a Transformer-driven Conditional Variational Autoencoder (CVAE). 
The model receives multimodal input consisting of follower joint states (angle, velocity, and torque), RGB images from both overhead and gripper-mounted cameras, and natural language instructions. 
It outputs action sequences predicting the leader robot’s joint trajectories, including angles, velocities, and torques.

Textual commands are encoded using SigLIP, which converts language instructions (e.g., ``put ball upward'') into fixed-length embeddings. 
These embeddings are fused with visual features extracted by EfficientNet through FiLM-based modulation, aligning language with visual context. 
The fused vision-language features, combined with robot joint angle, velocity, and torque data, form a unified latent space. 
This latent representation is then processed by a CVAE-based Transformer encoder-decoder, which generates action chunks supervised by minimizing reconstruction error with respect to ground-truth leader trajectories collected during bilateral control demonstrations.

By jointly embedding robot data, vision, and language, Bi-VLA enables a single policy to adaptively generalize across multiple tasks, addressing the core limitation of prior bilateral control-based imitation learning methods that required separate policies for each task.

\subsection{Inference}\label{sec:bi-vla_inference}
During inference, as shown in Fig.~\ref{fig:bi-vla_inference}, the Bi-VLA model receives the most recent follower joint states, synchronized camera images, and natural language instructions specifying the intended action. 
The model predicts the next action chunk of the leader robot joint trajectories, including angle, velocity, and torque. 
These outputs are converted into current commands by the bilateral control system and applied to the follower robot in real time. 
This closed-loop execution enables Bi-VLA to generate task-appropriate actions conditioned on multimodal inputs, supporting flexible, vision-language driven manipulation in real environments.

\section{Experiments}
\subsection{Overview}
We evaluate the proposed Bi-VLA, a policy architecture that integrates language and vision for robot imitation learning, through real-world pick-and-place experiments, as shown in Fig.~\ref{fig:ex1_tele}-\ref{fig:ex2_tele}.
To this end, we designed tasks that require either language-based or vision-based disambiguation, allowing us to probe the complementary roles of linguistic grounding and visual perception.
All experiments were conducted using a bilateral teleoperation setup with two manipulators: a leader robot, controlled by a human operator to provide demonstrations, and a follower robot, which executed policies trained on these demonstrations.
This setup enables a systematic comparison of language-dependent and vision-dependent tasks, thereby validating the effectiveness of Bi-VLA in multimodal robotic learning.
\begin{figure}[t] 
    \centering \includegraphics[keepaspectratio, width=0.98\linewidth]{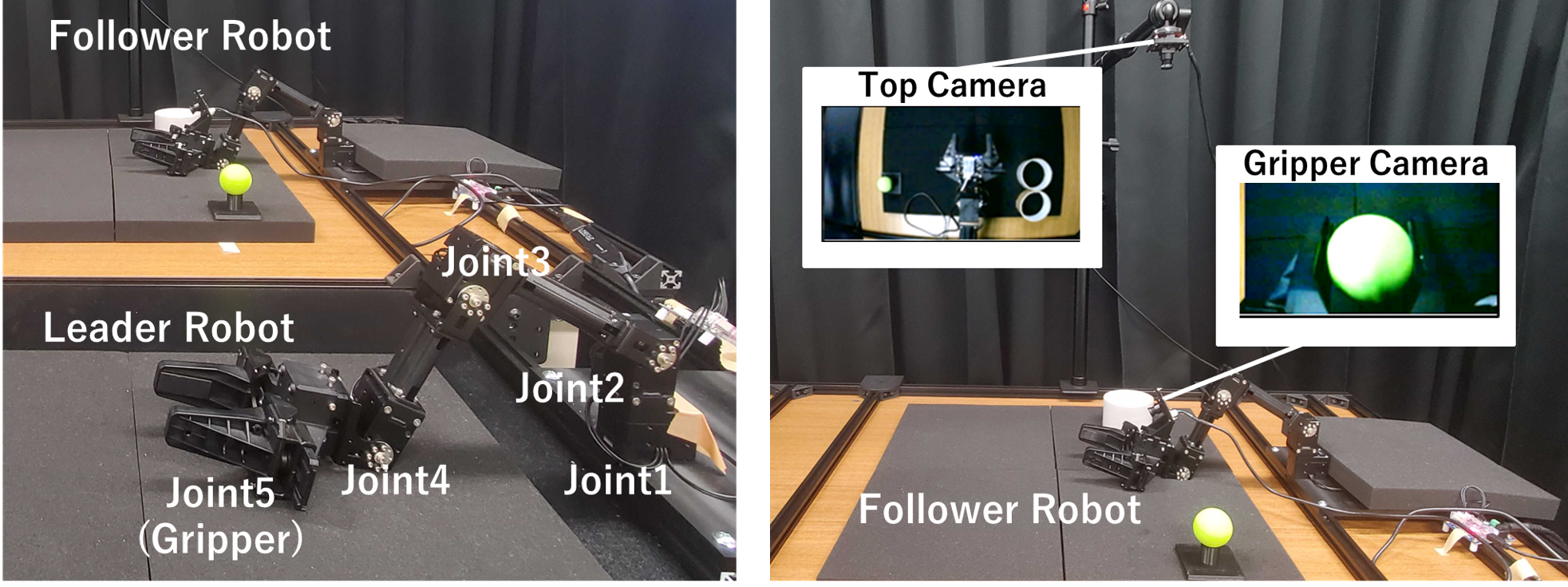} 
    \caption{ Experimental Environments} 
    \label{fig:ex_hardware} 
\end{figure}
\subsection{Hardware Setup}

\begin{table*}[t]
\centering
\caption{Training Configuration of Evaluated Models}
\label{tab:training_setup}
\begin{tabular}{l|c|c|c}
\hline
Model & Training Scope & Language Encoder & Demonstrations \\
\hline
Bi-ACT & Two-Target or Two-Source (separate data per model) & None & 6 (3 Up, 3 Down) \\
Bi-VLA (DistilBERT) & Two-Target only & DistilBERT & 6 (3 Up,  3Down) \\
Bi-VLA (SigLIP) & Two-Target or Two-Source (separate data per model) & SigLIP & 6 (3 Up, 3 Down) \\
Bi-VLA (SigLIP-Mix) & Two-Target and Two-Source (mixed data for model)& SigLIP & 4 (Two-Target\&Source 1 UP, 1 Down ) \\
\hline
\end{tabular}
\end{table*}
Fig.~\ref{fig:ex_hardware} shows the hardware and experimental environments.
We used the OpenManipulator-X from ROBOTIS.
The OpenManipulator-X arm is equipped with four rotational joints (Joint1-Joint4) and an additional fifth joint that operates the gripper.
In experiments, two robots were used: one configured as the leader robot and the other as the follower robot.
RGB images were recorded using an ELP USB device (model: ELP-USBFHD08S-L36), providing frames at 640×360 resolution.
Two of these cameras were deployed overhead and on the gripper to ensure comprehensive visual coverage.

\subsection{Task Setup}
We consider two pick-and-place tasks designed to evaluate language disambiguation and visual disambiguation. We intentionally present the language-disambiguable task first to highlight Bi-VLA's core capability.

\subsubsection{Two-Target Task (Language-Disambiguable)} 
A ball is picked from a fixed source and placed at one of two targets (Up/Down) specified by a language command, as shown in Fig.~\ref{fig:ex1_tele}.
Since the initial visual observation is identical across targets, vision alone cannot disambiguate the goal; the ambiguity must be resolved through language. 

\subsubsection{Two-Source Task (Vision-Disambiguable)}
The ball is picked from one of two source locations (Up/Down), and must be placed at a fixed target, as shown in Fig.~\ref{fig:ex2_tele}.
Here, vision alone is sufficient to infer the correct pick location. 
We also evaluated unlearned 3-ball environments to partially degrade visual saliency.

For each case, we conducted $n{=}10$ independent evaluation trials.
A trial is counted as \textit{success} only if all phases (Pick, Move, Place) are completed without unintended object drops outside the target area.

\subsection{Training Setup}
\begin{figure}[t] 
    \centering \includegraphics[keepaspectratio, width=0.98\linewidth]{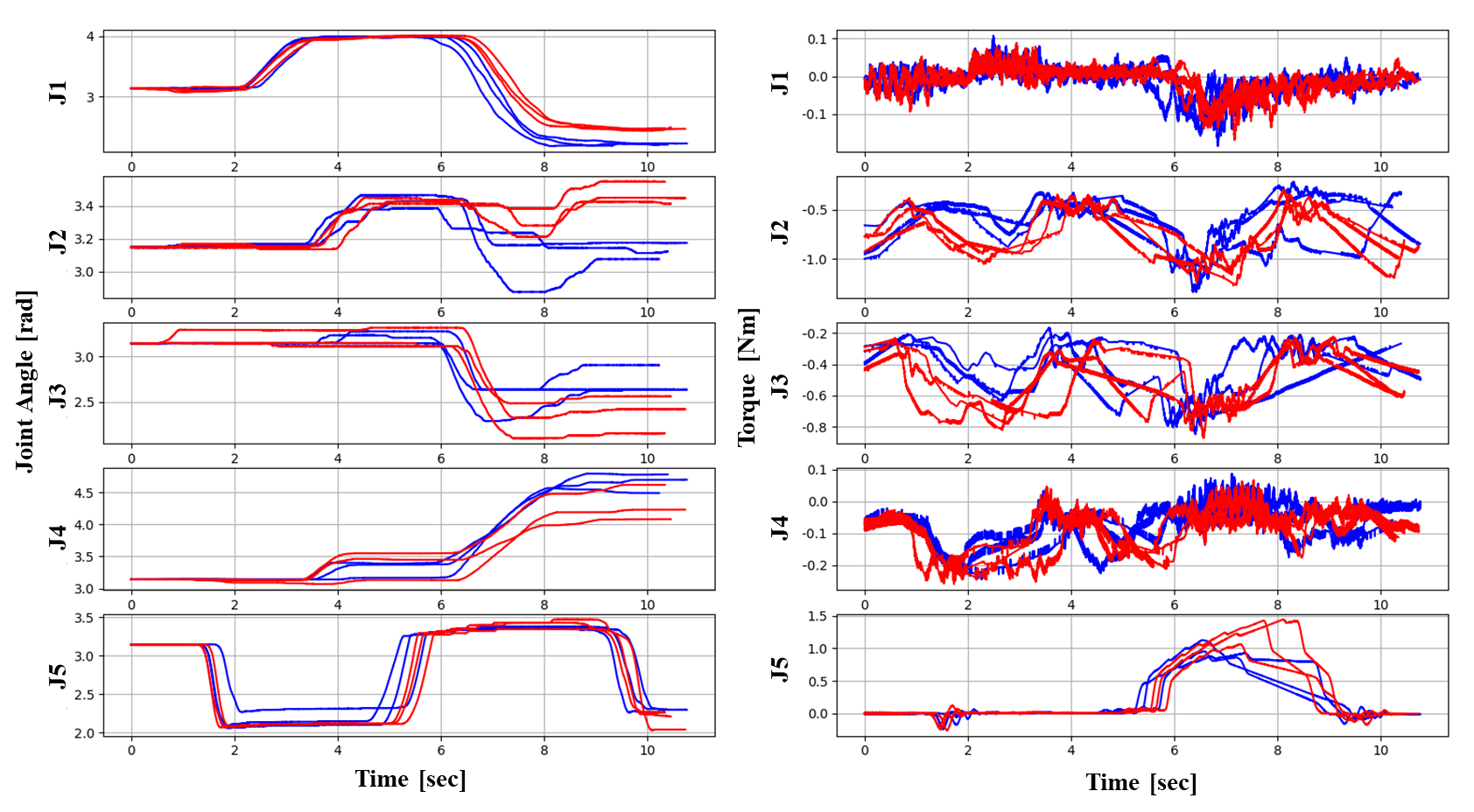} 
    \caption{ Two-Target Joint Data (Red: UP, Blue: Down)} 
    \label{fig:data_t} 
\end{figure}
\begin{figure}[t] 
    \centering \includegraphics[keepaspectratio, width=0.98\linewidth]{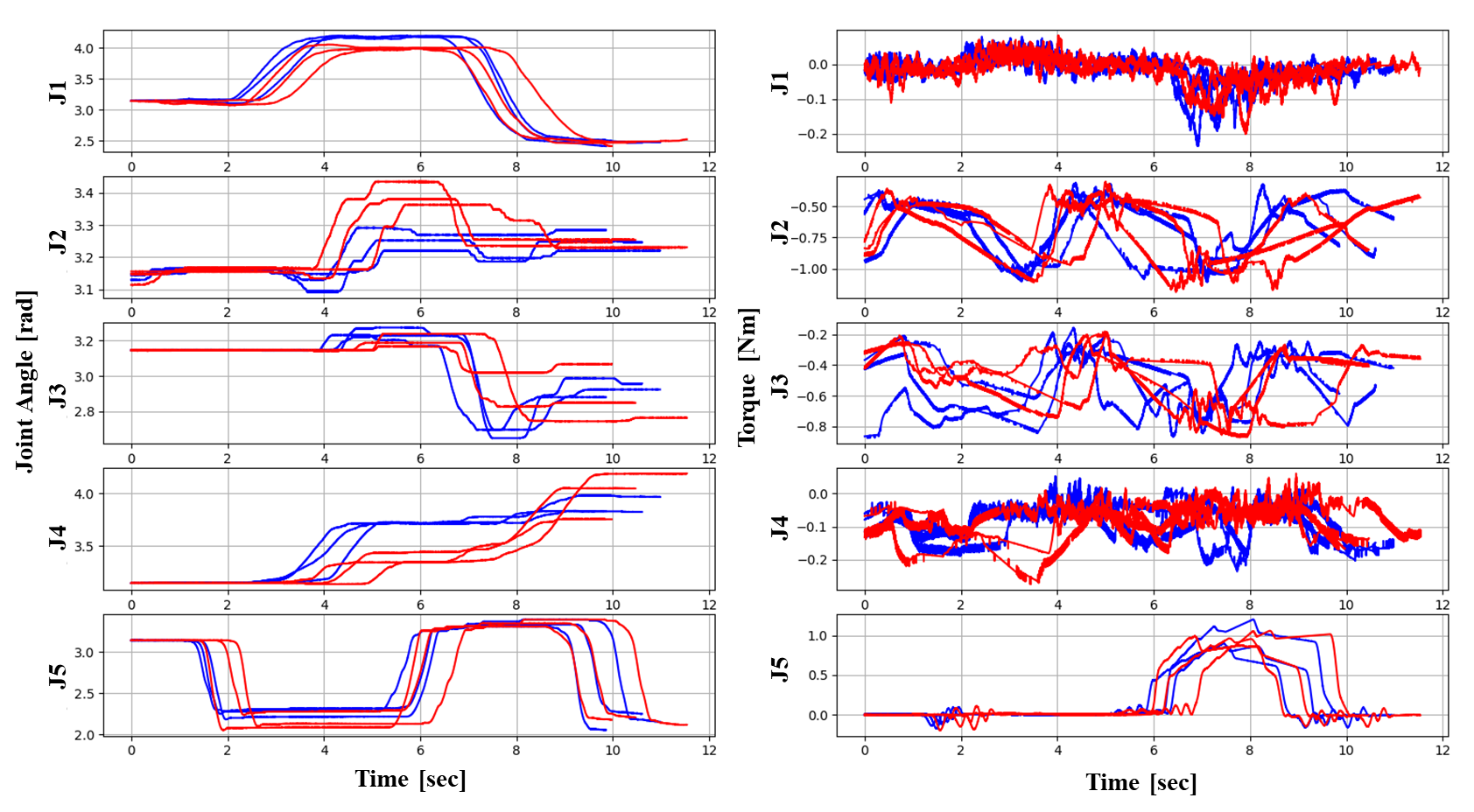} 
    \caption{ Two-Source Joint Data (Red: UP, Blue: Down)} 
    \label{fig:data_s} 
\end{figure}
\begin{figure*}[t] 
    \centering 
    \includegraphics[keepaspectratio, width=0.92\linewidth]{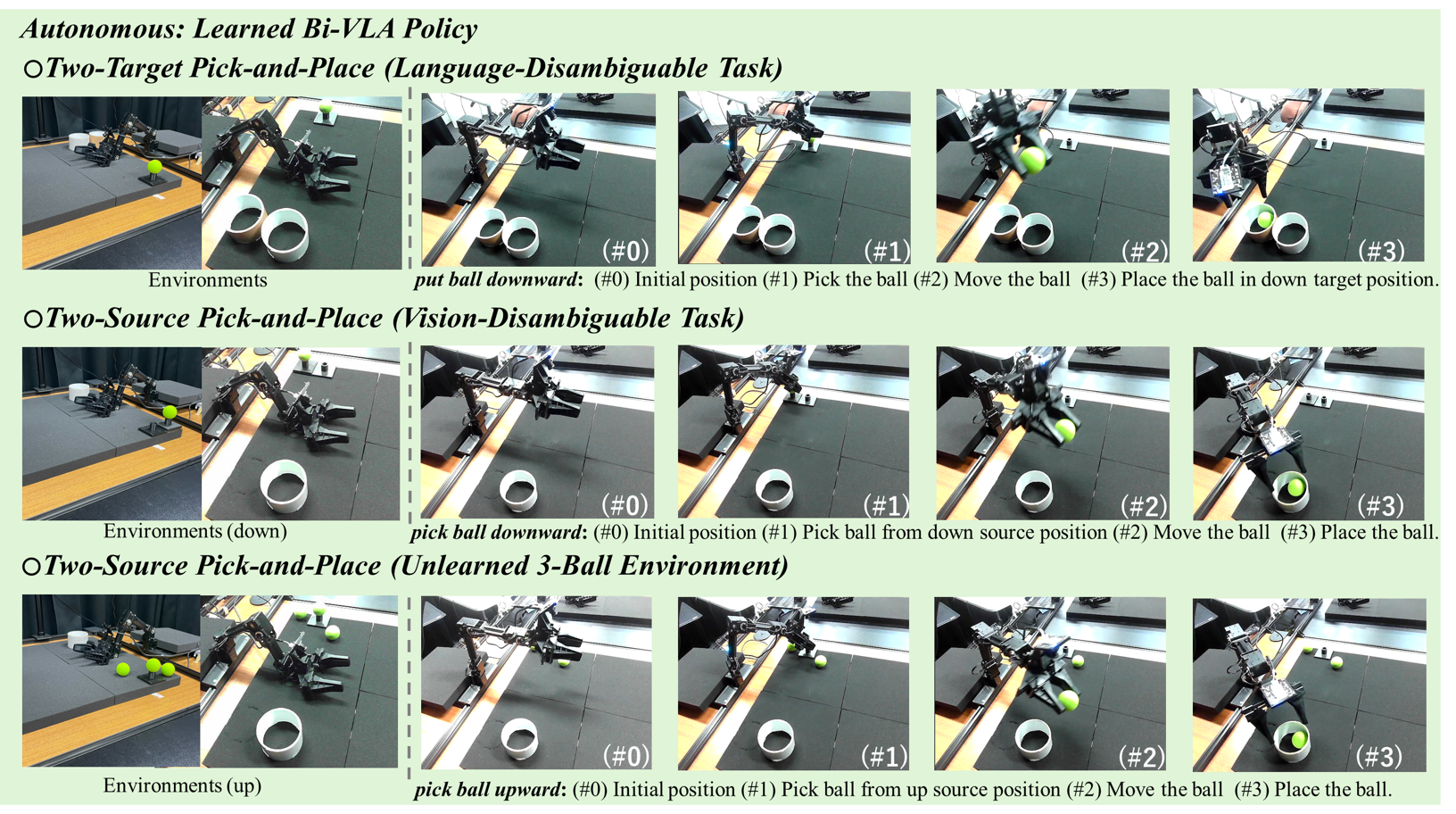} 
    \caption{Results of Experimental Task using Bi-VLA} 
    \label{fig:ex_auto} 
\end{figure*}

Using a four-channel bilateral control setup running at 1000~Hz, we logged the joint positions, velocities, and torque values of both leader and follower manipulators.
Each robot contributed 15 features (5 joints × 3 measurements), producing a combined 30-dimensional state vector that reflects fine-grained motor dynamics crucial for imitation learning.
Simultaneously, RGB images were captured from two cameras at a frequency of 100~Hz.

For each task, we collected six demonstrations, consisting of three trials for the Up condition and three for the Down condition, as shown in Fig.\ref{fig:data_t}-\ref{fig:data_s}
Each demonstration was paired with a natural language instruction:
\begin{itemize}
    \item Two-Target: ``put ball upward'' / ``put ball downward''
    \item Two-Source: ``pick ball upward'' / ``pick ball downward''
\end{itemize}

We evaluated four policy variants:
\begin{enumerate}
    \item Bi-ACT: vision-action baseline without language input.
    \item Bi-VLA (DistilBERT): trained only on the Two-Target task, to examine the effect of a conventional language encoder.
    \item Bi-VLA (SigLIP): trained with SigLIP as the language encoder, using Up/Down demonstrations.
    \item Bi-VLA (SigLIP-Mix): multitask training on both Two-Target and Two-Source tasks under reduced demonstrations.
\end{enumerate}

In the Two-Target task, Bi-VLA was trained with two different language encoders (DistilBERT and SigLIP) to examine the role of language representation.
Bi-ACT, which does not incorporate any language encoder, served as a vision--action baseline.

Moreover, to augment the training data, we employed the DABI~\cite{kobayashi2024dabievaluationdataaugmentation}.
The DABI approach is especially advantageous in scenarios where fast sensor streams from the robot must be synchronized with slower image sequences.
After reducing the 1000~Hz control signals to 100~Hz, DABI augmented the dataset from 6 demonstrations to 60, effectively enlarging it tenfold.
Further methodological specifics can be found in DABI~\cite{kobayashi2024dabievaluationdataaugmentation}.
Based on these datasets, we trained Bi-ACT, Bi-VLA(DistilBERT), and Bi-VLA (SigLIP).

For Bi-VLA (SigLIP-Mix), we deliberately imposed a reduced data budget to test cross-task generalization.
Only four raw demonstrations were collected in total—one for each condition (Target-Up, Target-Down, Source-Up, Source-Down).
After DABI augmentation, this produced 40 training data.
This design provided a controlled setting to evaluate whether a single multimodal policy could accommodate heterogeneous tasks without task-specific retraining.

We configured Bi-ACT and all Bi-VLA models with 4 encoder layers and 7 decoder layers.

\subsection{Experiment Results}

\subsubsection{Two-Target Task (Language-Disambiguable)}
Table~\ref{tab:two_target} and Fig.~\ref{fig:ex_auto} present the results for the language-disambiguable task, where the ball must be placed at either the ``Up'' or ``Down'' target according to a language command. 
Without access to language, Bi-ACT completely fails to generalize across both commands: it succeeds in all ``Up'' trials (100\%) but collapses to an ``Up-only'' policy, yielding 0\% success in the ``Down'' case. 
This biased behavior results in an overall success rate of only 50\%.  

When Bi-VLA used DistilBERT as the language encoder, performance improved modestly. The model achieves perfect success in the ``Up'' direction (100\%) and also consistently identifies the correct ``Down'' target during the pick and move phases (100\%). 
However, it fails to reliably complete the placement step in the ``Down'' case, with success dropping to 20\%. This bottleneck reduces the overall success to 60\%.  

In contrast, Bi-VLA with SigLIP demonstrates robust performance across both commands. 
Although performance in the ``Up'' case is slightly reduced to 80\%, the model achieves perfect success across all stages in the ``Down'' case (100\%). 
Taken together, these results in the highest overall success rate of 90\%.  

Finally, Bi-VLA (SigLIP-Mix) achieves a balanced success rate of 70\% across both ``Up'' and ``Down'' conditions. While this is lower than the task-specific SigLIP model (90\%), it is markedly higher than Bi-ACT (50\%) and comparable to Bi-VLA (DistilBERT) (60\%). Importantly, unlike Bi-ACT, which collapses into a biased ``Up-only'' strategy, the SigLIP-Mix variant successfully handles both command directions, underscoring its ability to incorporate linguistic cues even under limited supervision.

These findings highlight two important insights: (i) language input is indispensable when vision alone cannot disambiguate the goal, and (ii) SigLIP achieves higher grounding accuracy and execution reliability than DistilBERT, suggesting that different encoders yield markedly different levels of task performance.
\begin{table}[t]
    \centering
    \caption{Two-Target (Language-Disambiguable)}
    \label{tab:two_target}
    \begin{tabular}{l|c|cccc}
    \hline
    Method & Target & Pick & Move & Place & Overall \\
    \hline
    Bi-ACT & Up   & 100 & 100 & 100 & 100 \\
                         & Down & 0   & 0   & 0   & 0   \\
                         &      &     &     &     & \underline{50} \\
    \hline
    Bi-VLA (DistilBERT)  & Up   & 100 & 100 & 100 & 100 \\
                         & Down & 100 & 100 & 20  & 20  \\
                         &      &     &     &     & \underline{60} \\
    \hline
    Bi-VLA (SigLIP) & Up & 80  & 80  & 80  & 80  \\
                         & Down & 100 & 100 & 100 & 100 \\
                         &      &     &     &     & \underline{90} \\
    \hline
    Bi-VLA (SigLIP-Mix) & Up & 70 & 70 & 70 & 70 \\
                        & Down & 70 & 70 & 70 & 70 \\
                        &      &     &     &     & \underline{70} \\
    \hline
    \end{tabular}
\end{table}

\subsubsection{Two-Source Task (Vision-Disambiguable)}
\begin{table}[t]
    \centering
    \caption{Two-Source (Vision-Disambiguable)}
    \label{tab:two_source}
    \begin{tabular}{l|c|cccc}
    \hline
    Method & Source & Pick & Move & Place & Overall \\
    \hline
    Bi-ACT & Up   & 90 & 90 & 90 & 90 \\
                         & Down & 100 & 100 & 100 & 100 \\
                         &      &    &    &    & \underline{95} \\
    \hline
    Bi-VLA (SigLIP)        & Up   & 100 & 100 & 100 & 100 \\
                         & Down & 80  & 80  & 80  & 80  \\
                         &      &     &     &     & \underline{90} \\
    \hline
    Bi-VLA (SigLIP-Mix) & Up & 100 & 100 & 100 & 100 \\
                        & Down & 80 & 80 & 80 & 80 \\
                        &      &     &     &     & \underline{90} \\
    \hline
    \end{tabular}
\end{table}

The results for the vision-disambiguable task are presented in Table~\ref{tab:two_source} and Fig.~\ref{fig:ex_auto}. In this setting, the ball is placed at one of two distinct source locations, such that visual perception alone provides sufficient cues for disambiguation.  

For Bi-ACT, performance is highly stable across both source conditions: 90\% success in the ``Up'' case and 100\% in the ``Down'' case, yielding an overall average of 95\%.  

For Bi-VLA (SigLIP), the policy achieves perfect accuracy in the ``Up'' case (100\%), but its success rate drops to 80\% in the ``Down'' case, resulting in an overall rate of 90\%. 
Although this asymmetry indicates some sensitivity to spatial variation, the aggregated performance remains comparable to the baseline.  

Bi-VLA (SigLIP-Mix) maintains 100\% success in the ``Up'' case and 80\% in the ``Down'' case, yielding an overall rate of 90\%. This is nearly identical to the task-specific SigLIP model, suggesting that multitask training does not hinder vision-based disambiguation.

Importantly, these findings serve as a control case: when the task is fully disambiguable by vision alone, explicit language grounding does not provide additional benefits, but neither does it hinder performance. In other words, the integration of language grounding does not introduce any negative interference when unnecessary. Both policies are able to exploit visual cues alone to achieve reliable success.
The advantage of Bi-VLA (SigLIP-Mix) becomes clearer in subsequent experiments, where visual information is insufficient and language grounding is indispensable for robust generalization.

\subsubsection{Two-Source Task (Unlearned 3-Ball Environment)}
Table~\ref{tab:two_source_obst} and Fig.~\ref{fig:ex_auto} report the results for the more challenging generalization test, where policies trained on the two-source task were evaluated in an unlearned 3-ball environment with an additional distractor.  

Bi-ACT again collapses to a biased policy, succeeding in all ``Up'' trials (100\%) but completely failing in the ``Down'' case (0\%). 
This asymmetric behavior results in an overall success rate of only 50\%.  

Bi-VLA (SigLIP), by contrast, maintains robust performance. The policy achieves 100\% success in the ``Up'' case and 50\% success in the ``Down'' case, yielding an overall rate of 75\%. 

Interestingly, Bi-VLA (SigLIP-Mix) achieves 90\% success in the ``Up'' case and 60\% in the ``Down'' case, averaging 75\%. While this matches the task-specific SigLIP model in overall accuracy, the performance is more balanced between Up and Down.

These results suggest that incorporating language grounding enables policies to generalize more effectively to unlearned environments, even when distractors partially compromise visual saliency.
\begin{table}[t]
    \centering
    \caption{Two-Source (Unlearned 3-Ball Environment)}
    \label{tab:two_source_obst}
    \begin{tabular}{l|c|cccc}
    \hline
    Method & Source & Pick & Move & Place & Overall \\
    \hline
    Bi-ACT & Up   & 100 & 100 & 100 & 100 \\
                         & Down & 0   & 0   & 0   & 0   \\
                         &      &     &     &     & \underline{50} \\
    \hline
    Bi-VLA (SigLIP)        & Up   & 100 & 100 & 100 & 100 \\
                         & Down & 50  & 50  & 50  & 50  \\
                         &      &     &     &     & \underline{75} \\
    \hline
    Bi-VLA (SigLIP-Mix) & Up & 90 & 90 & 90 & 90 \\
                        & Down & 70 & 60 & 60 & 60 \\
                        &      &     &     &     & \underline{75} \\
    \hline
    \end{tabular}
\end{table}
\subsection{Summary and Discussion}
Across all evaluations (Tables~\ref{tab:two_target}--\ref{tab:two_source_obst}, Fig.~\ref{fig:ex_auto}), two consistent findings emerge. 
First, Bi-VLA (SigLIP) demonstrates reliable resolution of Up/Down ambiguity in language-dependent settings, achieving near-perfect execution for the ``Down'' command while preserving robust performance in the ``Up'' case. Unlike the baseline vision--action model, which degenerates into an ``Up-only'' strategy, the SigLIP-based variant consistently grounds linguistic instructions into appropriate action sequences, confirming the critical role of accurate language--vision alignment when visual cues alone are insufficient.

Second, Bi-VLA (SigLIP-Mix) exhibits strong multitask and low-data generalization. Trained jointly on Two-Target and Two-Source tasks with only four raw demonstrations and DABI augmentation, it maintains balanced success rates in the language-disambiguable task, matches task-specific SigLIP in the vision-disambiguable setting, and achieves competitive performance under unlearned 3-ball conditions. Importantly, this is accomplished without negative interference between tasks, indicating that a single policy can flexibly adapt to heterogeneous scenarios under limited supervision.

Taken together, these results highlight the strengths of Bi-VLA: high-fidelity grounding of linguistic instructions, robustness in vision-based tasks, and the ability to generalize across multiple tasks with minimal data. These properties position Bi-VLA as a promising framework for scalable multimodal imitation learning in real-world robotic applications.

\section{CONCLUSIONS}\label{sec:AW}
In this paper, we proposed Bi-VLA, a bilateral control-based imitation learning that integrates robot joint angle, velocity, and torque data with visual and language modalities. 
Unlike prior bilateral approaches such as Bi-ACT, which were limited to single-task execution, Bi-VLA enabled a single policy to perform multiple tasks by leveraging FiLM-based vision-language fusion and SigLIP-based language embeddings.
Overall, these results provided empirical evidence that combining vision and language within a bilateral control-based imitation learning framework enables flexible task switching under a unified model, thereby overcoming the single-task limitation of prior methods.
Real-robot experiments confirmed its effectiveness in both vision- and language-dependent scenarios.

While this paper demonstrated the effectiveness of Bi-VLA, the evaluation was limited to a small set of tasks and environments. Future work will extend the framework to more complex multi-step manipulations and diverse object categories, as well as validate its performance across different robotic platforms.
In addition, integrating larger-scale datasets and self-supervised learning could further enhance the generalization and real-world applicability of Bi-VLA.


\bibliographystyle{IEEEtran}

\end{document}